%% file: main.tex
\documentclass[10pt,twocolumn,letterpaper]{article}
\usepackage[pagenumbers]{cvpr} 

\definecolor{cvprblue}{rgb}{0.21,0.49,0.74}
\usepackage[pagebackref,breaklinks,colorlinks,allcolors=cvprblue]{hyperref}
\usepackage{mathrsfs}
\input{packages}

\input{macros}

\def\confName{CVPR}
\def\confYear{2025}
\title{In-Context Brush: Zero-shot Customized Subject Insertion \\ with Context-Aware Latent Space Manipulation}

\author{
Yu Xu\textsuperscript{1,2},
Fan Tang\textsuperscript{1,2},
You Wu\textsuperscript{1,2},
Lin Gao\textsuperscript{1,2},
Oliver Deussen\textsuperscript{3}, 
Hongbin Yan\textsuperscript{2}, \\
Jintao Li\textsuperscript{1}, 
Juan Cao\textsuperscript{1,2},
Tong-Yee Lee\textsuperscript{4}
}

\begin{document}
\twocolumn[{%
\renewcommand\twocolumn[1][]{#1}%
\maketitle


\begin{center}
    \captionsetup{type=figure}
    \includegraphics[width=\linewidth]{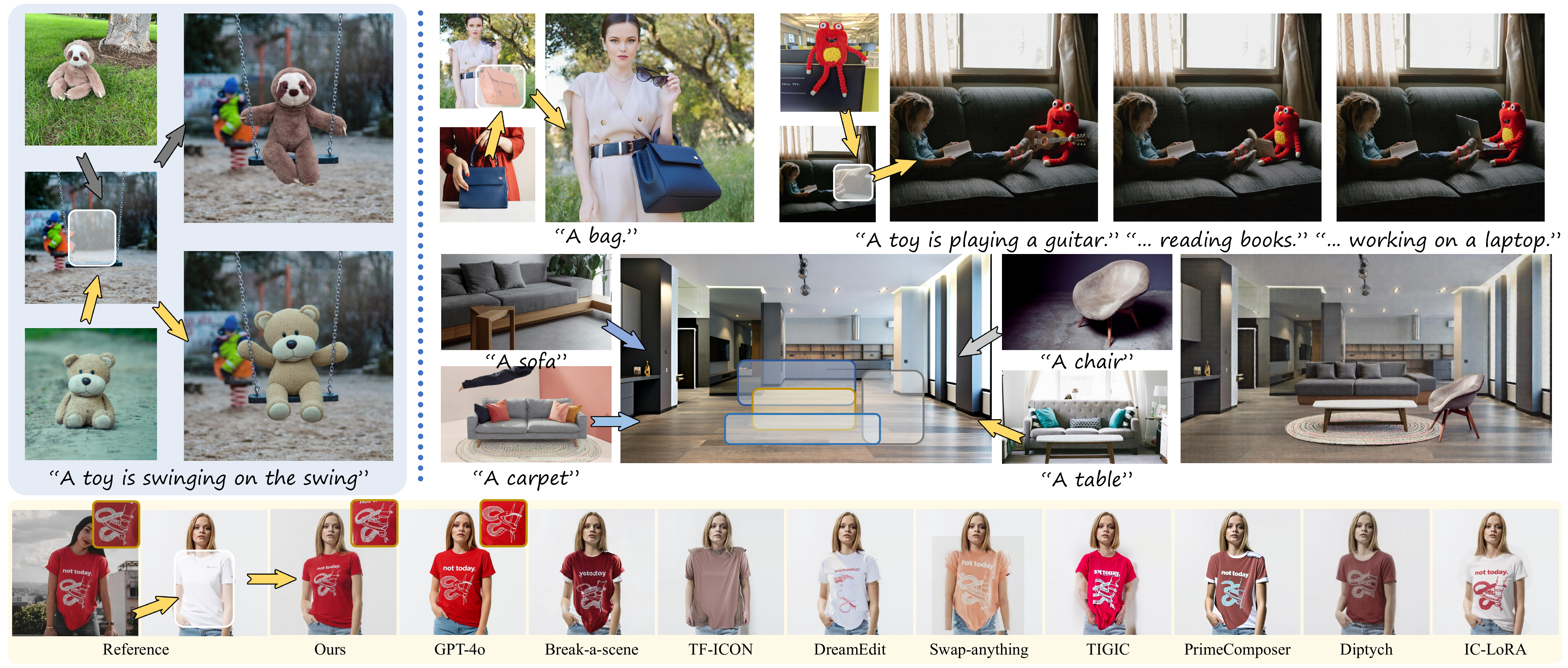}
    \vspace{-6mm}
    \caption{{Our method achieves identity-preserving subject insertion in the novel scene harmoniously, simultaneously enabling diverse text-driven control.}}
   \label{fig:teaser}
\end{center}
}]

\renewcommand{\thefootnote}{}
\footnotetext{{
$^{1}$Institute of Computing Technology, Chinese Academy of Sciences; $^{2}$University of Chinese Academy of Sciences; $^{3}$University of Konstanz; $^{4}$National Cheng-Kung University
}}

\input{sec/0_abstract}    
\input{sec/1_intro}

\input{sec/2_relate}
\input{sec/4_Method}

\input{sec/5_Experiments}

\input{sec/6_app}
\input{sec/7_limit_and_conclusion}

{
    \small
    \bibliographystyle{ieeenat_fullname}
    \bibliography{main}
}

\clearpage
\appendix
\input{sup}  

\end{document}

%% file: packages.tex
\usepackage{color, soul}
\usepackage{booktabs}
\usepackage{makecell}
\usepackage{verbatim}
\usepackage{tabularray}

\usepackage{multirow}
\usepackage{xcolor}
\usepackage{ifthen}
\usepackage{microtype}
\usepackage{amsmath}
\usepackage{amssymb}
\usepackage{graphicx}
\usepackage{amsfonts}
\usepackage{array}
\usepackage{stfloats}
\usepackage{url}
\usepackage{ragged2e}
\usepackage{ulem}
 
\hypersetup{colorlinks}
\usepackage{caption}
\usepackage{subcaption}

\usepackage{enumitem}
\usepackage{engord}
\usepackage{fmtcount}
\usepackage{microtype}
\usepackage{balance}
\usepackage{mathrsfs}
\usepackage[mathscr]{eucal}

\usepackage{enumitem}
\usepackage[ruled]{algorithm2e} 
\usepackage[super]{nth}

\usepackage{calligra}

\definecolor{revision}{rgb}{0,0,0}
\newcommand{\revision}[1]{{\color{revision} #1}}

\usepackage{multirow}
\renewcommand{\thefootnote}{}

%% file: macros.tex
\definecolor{FanColor}{rgb}{0.8,0,0.8}

\definecolor{YuxinColor}{rgb}{0.54,0.18,0.88}

\newcommand{\nothing}[1]{}

\newcommand{\gl}[1]{{\color{blue}            {}}}
\newcommand{\glc}[1]{{\color{orange}         {}}}

\definecolor{figred}{rgb}{1,0,0}
\definecolor{figgreen}{rgb}{0,0.6,0}
\definecolor{figblue}{rgb}{0,0,1}
\definecolor{figpink}{rgb}{1,0.63,0.63}

\renewcommand{\paragraph}[1]{\textbf{#1}}

%% file: sec/0_abstract.tex
\begin{abstract}
\label{sec:abstract}
Recent advances in diffusion models have enhanced multimodal-guided visual generation, enabling customized subject insertion that seamlessly ``brushes'' user-specified objects into a given image guided by textual prompts.
However, existing methods often struggle to insert customized subjects with high fidelity and align results with the user's intent through textual prompts. 
In this work, we propose \textbf{In-Context Brush}, a zero-shot framework for customized subject insertion by reformulating the task within the paradigm of in-context learning. 
Without loss of generality, we formulate the object image and the textual prompts as cross-modal demonstrations, and the target image with the masked region as the query.
The goal is to inpaint the target image with the subject aligning textual prompts without model tuning.
Building upon a pretrained MMDiT-based inpainting network, we perform test-time enhancement via dual-level latent space manipulation: 
intra-head \textit{latent feature shifting} within each attention head that dynamically shifts attention outputs to reflect the desired subject semantics and 
inter-head \textit{attention reweighting} across different heads that amplifies prompt controllability through differential attention prioritization.
Extensive experiments and applications demonstrate that our approach achieves superior identity preservation, text alignment, and image quality compared to existing state-of-the-art methods, without requiring dedicated training or additional data collection.
\end{abstract}

%% file: sec/1_intro.tex
\section{Introduction}
\label{sec:intro}

Image customization~\cite{ruiz2023dreambooth, galimage}, where users aim to render specific subjects into new contexts, has received increasing attention with the advancement of text-to-image diffusion models~\cite{rombach2022high, peebles2023scalable, podellsdxl,esser2024scaling}.
Beyond synthesizing new scenes from scratch, a more practical and challenging task is to insert a customized subject into a specific region of existing images.
This task requires maintaining high semantic fidelity to the customized subject, ensuring contextual harmony with the background, and enabling flexible contextual adaptation (e.g., varying pose, attributes, interactions) with textual prompts provided by users. 

Initial attempts~\cite{yang2023paint, song2023objectstitch, chen2024anydoor, chen2024zero, song2024imprint} for customized subject insertion typically replace text prompts with subject embeddings, allowing visual specification of the subject but inherently limiting the generation under textual guidance.
Later efforts~\cite{choi2023custom, lidreamedit, gu2024photoswap, gu2024swapanything} adopt a more straightforward way of learning subjects by fine-tuning the model, and then inserting them into target scenes via additional editing modules. 
However, such a workflow suffers from subject overfitting and reduced editing controllability. 
Recent approaches~\cite{wang2024primecomposer, li2025tuning} share the core objective using techniques, such as inversion and blending, to learn and insert subjects in a training-free manner.
However, the low-dimensional latent representations derived from inversion processes inherently restrict textual control precision.
Achieving customized subject insertion that harmoniously integrates the subject with visual context (target images) while maintaining identity consistency and adhering to textual context (prompts) with a training-free framework remains challenging to be explored.

Large-scale pre-trained models~\cite{achiam2023gpt, touvron2023llama} demonstrate remarkable capabilities for context understanding and give rise to in-context learning (ICL)~\cite{brown2020language, min2022rethinking, dong2022survey}, a powerful paradigm that transfers knowledge and facilitates predictions by leveraging input-output pairs, termed as \textit{demonstrations~(demos)}, in a zero-shot manner. 
Similarly, Diffusion Transformers~(DiTs)~\cite{peebles2023scalable, chen2024pixartalpha, esser2024scaling, flux} present a promising avenue to incorporate ICL to enable controllable text-to-image generation by utilizing text-image pairs as \textit{demos} and vision/language conditions as \textit{queries}, generating images that incorporate information from demos while following the specified conditions~\cite{zeng2024can}.

However, current ICL-based image generation methods~\cite{wang2023images, wang2023context, najdenkoska2024context} primarily focus on shallow task adaptation of image-text correspondences in demonstration pairs (e.g., pixel-to-caption matching) while failing to disentangle and transfer abstract subject semantics (e.g., cross-demo categorical invariants or relational patterns).
Furthermore, these task-specific conditioning mechanisms conflate subject identity with environmental context, thereby constraining zero-shot generalization to novel subject-scene combinations.
As a result, directly leveraging existing ICL frameworks for customized image editing remains a significant challenge.


In this paper, we dive into the ICL framework to enable zero-shot subject insertion. 
Subject images and textual prompts serve as \textit{demos}, while target images act as \textit{queries} for conducting regional insertion. 
Following the ICL paradigm, where demos and queries are concatenated as input, we also concatenate prompt tokens and subject image tokens with target image tokens in DiTs to construct an ICL-based inpainting framework.
With this framework, we formulate fine-grained subject-level transfer as shifting hidden states in DiTs and propose an \revision{intra-head} \textit{latent feature shift} injection mechanism to incorporate hidden states of subjects and textual prompts into queries. 
This enables customized subject-level injection, maintaining consistency between subject and output images while aligning with textual prompts.
\revision{Additionally, we introduce inter-head attention activation to improve textual control to subjects according to various prompts, and token blending to improve consistency between the inserted subject and the background. }
Experiments on benchmark datasets show that our method successfully inserts customized subjects into new scenes, enables diverse prompt-driven control, preserves subject fidelity, and achieves coherent visual integration.
Our contributions can be summarized as follows.
\begin{itemize}[leftmargin=10pt]
\item{We propose \textbf{In-Context Brush}, a zero-shot customized subject insertion framework that leverages ICL to transfer subject-level features in large-scale text-to-image diffusion models, and achieves superior identity preservation, prompt alignment, and image quality compared to state-of-the-art methods.}
\item{We reformulate subject insertion under the ICL paradigm as a latent feature shifting problem, and introduce a feature shift injection mechanism to enable accurate and consistent transfer of subject semantics into target scenes.}
\item{We further introduce attention head activation for prompts expressiveness enhancement, and propose a token blending strategy to ensure visual coherence between the inserted subject and the surrounding context.}
\end{itemize}

%% file: sec/2_relate.tex
\section{Related Work}
\label{sec:relate}

\input{Figures/pipeline}

\subsection{In-context learning for image generation}
With the scaling of model and dataset sizes, large language models (LLMs)~\cite{devlin2018bert, radford2019language, achiam2023gpt, touvron2023llama} have demonstrated remarkable ICL capabilities~\cite{brown2020language, wei2022emergent}. 
ICL enables models to learn from contextual demonstrations and apply the extracted knowledge to queries. This approach facilitates task execution by conditioning on a combination of demonstrations and query inputs, eliminating the need for parameter optimization. In recent years, the use of ICL has extended beyond natural language processing to encompass image generation.
Prompt Diffusion~\cite{wang2023context} introduces a framework that employs in-context prompts for training across various vision-language tasks, enabling the generation of images from vision-language prompts. Building on this, iPromptDiff~\cite{chen2023improving} enhances visual comprehension in visual ICL by decoupling the processing of visual context and image queries while modulating the textual input using integrated context.
Furthermore, Context Diffusion~\cite{najdenkoska2024context} separates the encoding of visual context from the query image structure, enabling the model to effectively leverage both visual context and text prompts.
However, previous works primarily focus on learning task relationships from demonstrations and transferring them to queries.
In contrast, our approach emphasizes learning the semantic feature relationships between subject and target images, enabling subject features insertion into specific regions through a training-free mechanism.

\subsection{Customized subject insertion with diffusion models}
Previous methods~\cite{yang2023paint, chen2024anydoor, song2024imprint, song2023objectstitch, chen2024zero, yu2025omnipaint} typically encode the subject image into embeddings that serve as input conditions to diffusion models. 
However, text conditions are replaced by image embeddings in the models, making it hard to guide the generation process with prompts. As a result, the output often does not match the user’s intended description, reducing its usefulness.
Recent zero-shot approaches~\cite{winter2024objectmate, song2025insert} construct large-scale datasets to train subject insertion models. However, the prompts used in training are typically limited to task-level instructions (e.g., object replacement or removal) or coarse descriptions of the entire scene, which restricts the ability to perform fine-grained control over the inserted subject.
Two-stage approaches~\cite{gu2024photoswap, gu2024swapanything, avrahami2023break, lidreamedit, zhuinstantswap} first learn subject-specific embeddings through customization techniques~\cite{ruiz2023dreambooth, galimage}, and then perform insertion into target scenes. While enabling prompt-driven editing, it comes at the cost of subject-specific training, reducing applicability in real-world scenarios.
Recently, training-free methods~\cite{wang2024primecomposer, lu2023tf, li2025tuning} have emerged to avoid tuning. These approaches perform inversion of both the subject and scene images into the diffusion latent space, then combine them via a training-free mechanism. 
However, these methods provide limited controllability through textual prompts due to the lack of explicit alignment between subject semantics and prompt guidance.
StyleAligned~\cite{hertz2024style} and ConsiStory~\cite{tewel2024training} explore the feature sharing between reference and target images for stylization and consistency generation tasks, While StyleAligned focuses on stylistic control, it lacks structural precision, limiting its use for subject insertion. ConsiStory ensures image consistency but struggles to preserve identity when learning from given images.
IC-LoRA~\cite{huang2024Incontextlora} activates the in-context generation capabilities of DiTs by training task-specific LoRA modules using paired datasets, but its reliance on data collection and retraining limits practicality. In contrast, our method is training-free and uses ICL to transfer subject features across tasks efficiently.
A concurrent work, Diptych Prompting~\cite{shin2024large} also leverages attention in a training-free manner. However, it re-weights attention to emphasize reference influence, which may overlook subject relationships and cause identity inconsistency. In contrast, our method integrates visual features and textual guidance via ICL, achieving stronger alignment with prompts while preserving subject identity.

%% file: Figures/pipeline.tex
\begin{figure*}[!t]
  \centering
  \includegraphics[width=1\linewidth]{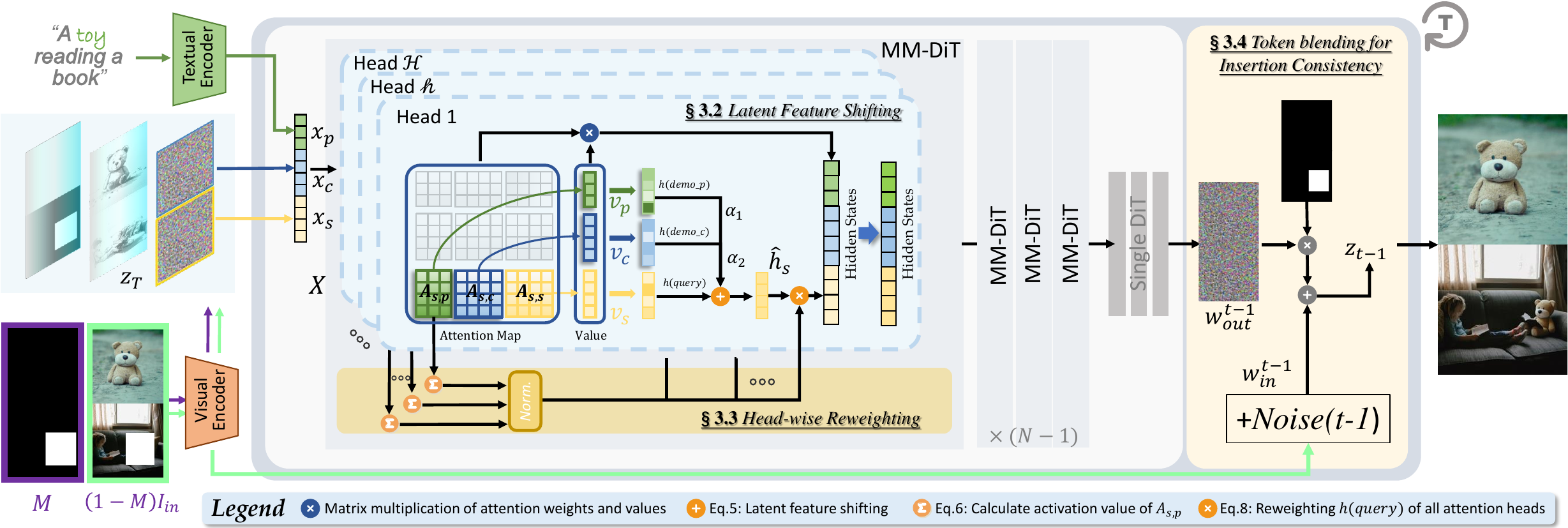}
  \caption{\textbf{Pipeline of our method.} We mainly introduce latent space shifting for subject present in target images in a training-free manner. In the ``Latent Feature Shifting'' part, features from the reference are shifted to output. We propose attention heads activation for further enhance representation of textual prompts and token blending for consistency injection within the image.}
  \vspace{-1.2em}
  \label{fig:pipeline}
\end{figure*}

%% file: sec/4_Method.tex
\section{Method}
\label{sec:method}

Given a subject image $I_{c} \in \mathbb{R}^{H \times W \times 3}$ containing the subject to be inserted, a target image $I_{s} \in \mathbb{R}^{H \times W \times 3}$ providing the background context, a textual prompt $p$ describing the desired output subject, and a binary mask $m \in \{0, 1\}^{H \times W}$ specifying the insertion region, we aim to transfer the subject from $I_c$ into the mask region of $I_s$ with guidance from user provided $p$, and get final output image $I_{gen}$.
To do so, in Sec.~\ref{3.1}, we formulate customized image insertion with ICL in DiTs.
In Sec.~\ref{3.2}, we introduce our core mechanism, latent feature shifting, which enables subject transfer in latent space.
In Sec.~\ref{3.3}, we present head-wise reweighting to enhance textual control.
In Sec.~\ref{3.4}, we describe token blending, which ensures better visual consistency between the inserted subject and the background.

\subsection{Preliminary}
\label{3.1}


\revision{We adopt multi-modal diffusion transformers (MM-DiTs)~\cite{esser2024scaling,flux} as the backbone of our generation framework. In each sampling step, MM-DiTs take a combination of text and image token embeddings as input and progressively denoise a latent representation to synthesize the output image.}
To integrate customized subject insertion into MM-DiTs, we introduce an ICL paradigm to model the subject-level relationships.

\revision{In the setting of ICL in LLMs, consider the translation task, given a few demo prompts, the model will infer task rules based on this task-wise contextual information and translate new input queries.
In our scenario, we propose a feature-wise ICL paradigm instead, which transfer subject feature from demo to query.}
Specifically, to utilize ICL, we construct input demonstrations analogous to those in large language models: the prompt $p$ and subject image $I_c$ jointly serve as \textit{demonstration (demo)}, providing contextual information, while the target image $I_s$ is the \textit{query} whose corresponding region will be inserted.
 Formally, we concatenate $I_c$ and $I_s$ into a single input image $I_{in} = \left[ I_{c}; I_{s} \right]$, and the mask is correspondingly extended as \revision{$M = \left[\mathbf{0}_{}; m\right]$}.

In this ICL-based configuration, MM-DiT implicitly learns to transfer subject-level features from the demo ($p$, $I_{c}$) into the query image ($I_{s}$) by latent space shifting, detailed in Sec.~\ref{3.2}.
To precisely insert the subject into the background image, we additionally apply Grounding DINO~\cite{liu2025groundingdino} and Segment Anything Model~(SAM)~\cite{kirillov2023SAM} to remove the original background in $I_{c}$, isolating the desired subject clearly.
As a result, with a generation model $G_{\theta}$, the output image $I_{gen} \in \mathbb{R}^{H \times W \times 3}$ can be formally predicted as:
\begin{equation}
  \begin{aligned} 
  \relax [I_{c}; I_{\text{gen}}] 
  &= G_{\theta}(p, I_{\text{in}}, M), \\
  &= G_{\theta} \big(p, [I_{c}; I_{s}], [\mathbf{0}_{H \times W}; m] \big). \\
  \end{aligned}
  \label{eq:formalization}
\end{equation}

\subsection{Latent feature shifting for subject injection}
\label{3.2}
In this section, we prove that subject-level features can be injected by shifting hidden states within the framework of ICL, effectively leveraging information from multi-modal demos.
In Sec.~\ref{3.1}, $p$ and $I_c$ are concatenated within attention blocks and used to compute the final hidden states through a joint-attention mechanism.
Specifically, let $X = \operatorname{Concatenate}(\left[x_{p}, x_{c}, x_{s}\right])$ represent the input embedding, where $x_{p}$, $x_{c}$ and $x_{s}$ represent input token embeddings at the same concatenating positions as $p$, $I_{c}$ and $I_{s}$, respectively.
Let $W_q$, $W_k$, and $W_v$ be the learnable key, query, and value matrices for computing the attention features $Q$, $K$, and $V$, the output hidden states of attention blocks can be formulated as:
\begin{equation}
  \begin{aligned} 
  \hat{h}=\operatorname{Attn}\left(X W_q, X W_k, X W_v\right)=\operatorname{Concatente}(\left[h_{p}, h_{c}, h_{s}\right]),
  \end{aligned}
  \label{eq:attention}
\end{equation}
where $h_{p}$, $h_{c}$, $h_{s}$ represent the hidden states corresponding to each component in $X$. We put the detailed derivation in supplementary materials.

Although in attention blocks, the overall feature $X$ is processed in a self-attention manner, there also exist relationships in the form of cross-attention among different pairs of its components.
For example, $h_{s}$ is directly composed of two parts: one part is derived from the self-attention computation of $x_{s}$; the other part is obtained through the interaction with features provided by the textual prompt and the reference subject, i.e., $x_p$ and $x_{c}$.
We only focus on $h_{s}$ because the generated result $I_{gen}$ is directly related to it.
This characteristic activates us to leverage contextual information from other features in the latent space from the perspective of in-context learning.
When $x_p$ and $x_{c}$ interact through cross-attention with $x_{s}$ respectively, they serve as demo providing semantic feature-wise contextual information and generating the attention output $h(demo\_p)$ and $h(demo\_c)$.  
As for the position for insertion, the self-attention computation of $x_{s}$ itself yields the original output $h(query)$ without demo.
Therefore, we rewrite the formula of $h_{s}$ in the form of the attention operation:

\small{
\begin{equation}
  \begin{aligned} 
  &h_{s}=\operatorname{Softmax}\left(\begin{bmatrix}
    x_{s} W_{qk} x_p^{\top} & x_{s} W_{qk} x_{c}^{\top} & x_{s} W_{qk} x_{s}^{\top}
    \end{bmatrix}\right) \begin{bmatrix}
    x_p W_v \\
    x_{c} W_v \\
    x_{s} W_v
    \end{bmatrix} \\
  &= \alpha_{p} \cdot h(demo\_p) + \alpha_{c} \cdot h(demo\_c) + \alpha_{s} \cdot h(query),
  \end{aligned}
  \label{eq:final}
\end{equation}
}
where $W_{qk} = W_{q} W_{k}^{\top}$.
\revision{We put the detailed derivation in supplementary materials.}
$\alpha_{tag}$ is the scalar that represents the sum of normalized attention weights between different hidden states:
{\scriptsize
\begin{equation}
  \begin{aligned} 
    \alpha_{tag}=\frac{\sum \exp \left(x_{s} W_{qk} x_{tag}^{\top}\right)}{\sum \exp \left(x_{s} W_{qk} x_{p}^{\top}\right)
    +\sum \exp \left(x_{s} W_{qk} x_{c}^{\top}\right) +\sum \exp \left(x_{s} W_{qk} x_{s}^{\top}\right)},
  \end{aligned}
  \label{eq:alpha_tag}
\end{equation}
}
where $\alpha_{p} + \alpha_{c} + \alpha_{s} =1$. 
Therefore, the essence of this subject-level relationship ICL can be regarded as a latent feature shifting on the original attention output $h(query)$ on the direction figured by $h(demo\_p)$ and $h(demo\_c)$.
The attention mechanism of DiTs automatically determines the distance of the shift.

Based on our conclusion, we propose a method named ``\textbf{feature shift injection}'', a straightforward way that manipulates the shift of attention feature outputs directly related to $I_{gen}$, to enhance the utilization and focus of DiTs on in-context information from input conditions in customized subject insertion.
Specifically, we can divide the weight map for each attention head within the attention blocks into multiple patches, as shown in Fig.~\ref{fig:pipeline}.

For the convenience of representation, we use $A_{i,j}$ to represent attention map in position of patch $x_i W_{qk} x_j^{\top}$, and define value feature $V = \operatorname{Concat}(\left[ v_p, v_c, v_s \right] = \operatorname{Concat}(\left[ x_p W_v, x_{c} W_v, x_{s} W_v \right]$.
The results of the hidden states $h_{s}$ are determined solely by the bottom three attention maps $A_{s,p}$, $A_{s,c}$, and $A_{s,s}$.
According to Eq.~\ref{eq:final}, they are respectively computed with the corresponding three parts of the value feature $V$ to obtain $h(demo\_p)$, $h(demo\_c)$, and $h(query)$.

To shift the latent features from $h(query)$, we directly amplify the values of scalars $\alpha_{p}$ and $\alpha_{c}$ because they are controlling the influence of $h(demo\_p)$ and $h(demo\_c)$ on the original latent feature $h(query)$ without demos.
In fact, this corresponds to adding the weighted results of separately computing attention maps $A_{s,p}$ and $A_{s,c}$ with $v_p$ and $v_c$ onto the output latent states $h_{s}$:
\begin{equation}
  \begin{aligned} 
    \hat{h}_{s} 
    &=h_{s} 
    + \alpha_{1} A_{s,p} v_p + \alpha_{2} A_{s,c} v_c,
  \end{aligned}
  \label{eq:ours}
\end{equation}
where $\alpha_{1}$ and $\alpha_{2}$ control the strength of shift like Eq.~\ref{eq:final}. 
\revision{Through the shifting operation within the ICL mechanism, we inject hidden states of \textit{demo}, which include the features of the subject and the textual prompt, to the output image, enabling capture the subject-level relationships from in-context conditions and generate consistent subjects in a training-free manner.
}

\subsection{Head-wise reweighting for textual control injection}
\label{3.3}
\revision{
While latent feature shifting mechanism enables subject transfer, effective control with diverse prompts remains challenging due to strong priors encoded in the reference image. 
In practice, we observe that inserted subjects often retain undesired attributes (e.g., colors, materials) from the subject image, even when the prompt specifies changes. This common limitation stems from the lack of selective control over semantic attention during generation.
To address this, we introduce a head-wise reweighting mechanism that improves the alignment between the generated image and the prompt by adaptively adjusting the contribution of different attention heads. This is motivated by recent findings~\cite{gandelsmaninterpreting, xu2024headrouter} that attention heads in transformers exhibit semantic specialization—different heads respond to different types of features.
Our key insight is to leverage the attention activation in the demo of the ICL setup to estimate which attention heads are most activated by the prompt tokens, and then reweight these heads during generation.
}
As shown in Eq.~\ref{eq:final}, we leverage $h(demo\_p)$ to soft activate $h(query)$ across different attention heads. 
Specifically, for $h(demo\_p)$, we measure the attention maps $A_{p,s}$ across all attention heads and assign different weights to queries based on activation values.
The activation value of $A_{p,s}$ in attention head $\mathsf{h}$ can be formed as: 
\begin{equation}
V_\mathsf{h} = \sum_{i,j} \left( A_{p,s}^{(\mathsf{h})} \right)_{i,j},
\end{equation}
where $i$ and $j$ are indices of $A_{p,s}^{(\mathsf{h})}$. Then we normalize all $V_{\mathsf{h}}$ across attention heads following:
\begin{equation}
\hat{V}_{\mathsf{h}} = \frac{V_{\mathsf{h}} - \min(V)}{\max(V) - \min(V)}, \mathsf{h} = 1,2,...\mathcal{H},
\end{equation}
where $min(V) = min\{V_{1}, V_{2},...,V_{\mathcal{H}}\}$,
$max(V) = max\{V_{1}, V_{2},...,V_{\mathcal{H}}\}$.
The final output of hidden states on each attention head are:
\begin{equation}
    \hat{h}_{\mathsf{h}}(query) = h_{\mathsf{h}}(query) \cdot \hat{V}_{\mathsf{h}}.
\end{equation}

This encourages the model to emphasize semantic that are relevant to the user prompt while suppressing prompt-irrelevant heads. This improves semantic controllability and leads to more faithful editing with respect to user intent.

\subsection{Token blending for insertion consistency}
\label{3.4}
In this section, we tackle the challenge of ensuring intra-image consistency when inserting customized subjects by refining feature interactions to mitigate distribution shifts. Specifically, as the subject is injected into a new contextual environment, we further analyze the challenge of ensuring intra-image consistency on customized subject injection.

Due to the semantic differences between the inserted subject and the background, the latent feature of the insertion region within mask $m$ in $I_s$ could be incongruous with the background~(i.e., $I_s \cdot (1-m)$).
$I_s$ is expected to guide the inserted subject in $x_s$ to be consistent with target regions in distribution.
However, affected by $x_c$ and $x_p$ after each sampling step, the semantic distribution of 
target region in $x_s$ deviates from the original input after each sampling step. 
In the next step, the distribution-biased $x_s$ will, in turn, provide erroneous guidance for the subject insertion of the target region~(i.e., $x_s \cdot m$) due to the interaction of contextual information in DiTs. 
Multiple sampling steps will gradually amplify this bias, resulting in an inharmonious fusion effect between the inserted subjects and the background in the result (such as irregular edges or differences in tone).

To prevent the deviation caused by inconsistent distribution in multi-step sampling, we propose effective token blending for insertion consistency.
\revision{Specifically, suppose the output hidden states of denoising step $t$ is $w^t$, we add noise to $(1-M) \cdot I_{in}$ to obtain $\text{w}_{in}^{t-1}$ after each step $t$ to $t-1$. 
Subsequently, we fuse $\text{w}_{in}^{t-1}$ with the output $\text{w}_{out}^{t-1}$ of the current step according to the mask $M$:
\begin{equation}
  \begin{aligned} 
  \text{w}_{\text{out}}^{t-1} = \text{w}_{\text{in}}^{t-1} + \text{M} \cdot \text{w}_{\text{out}}^{t-1}.
  \end{aligned}
  \label{eq:formalization_2}
\end{equation}
}
By this method, we ensure that in each step, the inserted region can be correctly guided by unbiased background semantics in distribution, thus enhancing the consistency between the inserted subjects and the context of the background in $I_{gen}$.

%% file: sec/5_Experiments.tex
\section{Experiments}
\label{sec:experiments}

In this section, we first introduce the experiment settings, and present qualitative and quantitative results in Sec.~\ref{sec:Qualitative} and Sec.~\ref{sec:Quantitative}. We further evaluate two-stage methods that combine subject insertion methods with customization methods or editing methods for a comprehensive comparison. Finally, we conduct an ablation study on hyperparameters and proposed modules. For more details on multiple seeds and time cost, please refer to  supplementary materials.
\input{Figures/main_compare}

\input{Tables/alignment}
\subsection{Experiments settings}
\paragraph{Baselines} We compare our method with eight state-of-the-art text and image-guided image generation methods, including training-based methods Break-a-scene~\cite{avrahami2023break}, Swap-anything \cite{gu2024swapanything}, DreamEdit~\cite{lidreamedit}, IC-LoRA~\cite{huang2024Incontextlora}, and training-free methods TF-ICON~\cite{lu2023tf}, TIGIC~\cite{li2025tuning}, PrimeComposer~\cite{wang2024primecomposer} and Diptych Prompting~\cite{shin2024large}. We also include two-stage methods, which combine MimicBrush~\cite{chen2024zero} with TurboEdit~\cite{deutch2024turboedit} (first insert subjects to target images then editing images) and combine Dreambooth~\cite{ruiz2023dreambooth} with Paint-by-example~\cite{yang2023paint} (first learn and edit subjects then inject to target images) for further comparison.

\paragraph{Datasets} 
We collect subject images from Dreambooth datasets~\cite{ruiz2023dreambooth}, which contains 30 subjects of 15 different classes, and collect 50 diverse scenes as target images from COCO dataset~\cite{lin2014microsoft}. 
We also collected 50 additional subject images and 80 scene images from the Internet to enable a more diverse and comprehensive evaluation. Thus, the evaluation dataset contains 100 subject images and 130 scene images.

\subsection{Qualitative comparisons}
\label{sec:Qualitative}
We present the visual results compared with customized subject insertion methods in Fig.~\ref{fig:main_compare}.
TF-ICON struggles to maintain semantic information from subject images (in the cases of ``backpack'', ``barn'' and ``teapot'') and has difficulty in editing aligning with textual prompts.
Break-a-scene has a good ability to follow prompt guidance in most cases. However, it lacks accurate expression of fine-grained features (in the cases of ``toy'' and ``teapot''), and also some obvious artifacts are presented between subject and background (``backpack'', ``barn'' and ``mosaic tile''), leading to overall disharmony.
Swap-anything fails to learn the semantic information of the subject, leading to the expression of the subject in the results being close to the copy-move effect and showing limited effects when editing complex subjects.
DreamEdit, TIGIC, and PrimeComposer also fail to accurately edit the subject in accordance with the given prompts, while IC-LoRA fails when handling complex prompts such as ``barn'', ``toy'', or ``mosaic tile''. 
PrimeComposer also lacks semantic learning ability and generates subjects in copy-move effects (``toy'' and ``mosaic tile'').
Diptych struggles to achieve generate results with high identity alignment with reference subjects (in the case of ``backpack'', ``barn'', and ``teapot''), or editing effects (in the case of ``backpack'', ``toy'', and ``teapot'').
Our approach achieves the best identity preservation and prompt-followed editing effects, surpassing the performance of baseline methods.

\subsection{Quantitative comparisons}
\label{sec:Quantitative}
Following previous methods~\cite{lu2023tf,song2024imprint,gu2024swapanything,lidreamedit}, we evaluate our method in three aspects: subject identity alignment between $I_{c}$ and $I_{gen}$, editing alignment between $p$ and $I_{gen}$, and overall image quality. We use DINO~\cite{liu2025groundingdino} and CLIP-I~\cite{ilharco_gabriel_2021_5143773} for subject identity alignment, including subject injection results and subject injection with editing results. We use CLIP-T~\cite{ilharco_gabriel_2021_5143773} to evaluate editing alignment. ``Injection'' evaluates identity alignment between reference and generated images. ``Editing'' evaluates prompt alignment. FID~\cite{heusel2017gans} is used for evaluating overall image quality.
We conduct quantitative comparisons in Tab.~\ref{tab:alignment}, displaying the evaluation index results and their standard deviations. 
As show in Tab.~\ref{tab:alignment}, our method outperforms baselines in subject identity alignment, editing alignment, and image quality.
Although the editing score (third column) is slightly lower than Break-a-scene and Swap-anything, likely due to their training-based methods that enable more free representation of semantics and are not limited to the reference, our method achieves stronger CLIP-I and CLIP-T scores, indicating higher robustness without requiring additional training.
For fine-grained evaluation of subject fidelity and background preservation, following DreamEdit~\cite{lidreamedit} and SwapAnything~\cite{gu2024swapanything}, we segment the generated results into subject and target regions. We then compute the similarity between generated and reference images using DINO and CLIP-I features for both regions. These metrics reflect the semantic consistency of the inserted subject and the integrity of the surrounding context. As shown in Tab.~\ref{tab:alignment_foreback}, our method achieves the highest scores across all four metrics, demonstrating superior subject fidelity preservation and background maintenance.

\input{Figures/user_study}

\input{Tables/alignment_foreback}

\paragraph{User study.} We conduct a user study to evaluate participant preferences on identity alignment, editing alignment, and overall image quality between our method and the baselines. A total of 65 participants~(33 female and 32 male, aged 14 to 55 years) participated in the survey, including 25 researchers specializing in computer graphics or computer vision. Each participant was asked to evaluate 35 cases, resulting in 6,825 votes. We present results in Fig.~\ref{fig:user_study}, and from these, we can see that our method achieves the best identity alignment preference. This indicates that the subject identity of the original image is most effectively preserved in the generated image, avoiding distortion. Furthermore, our method receives the highest editing alignment preference, indicating better prompt-driven customization than other approaches, achieving the customized effects desired by users. Overall, users also favor our results for their higher image quality and visual coherence.

\subsection{Comparison with two-stage methods}
Customized subject insertion can also be achieved through two-stage approaches: by utilizing advanced subject customization techniques~\cite{ruiz2023dreambooth, galimage} in the first stage for training custom subject representations and leveraging image-guided editing methods~\cite{yang2023paint} in the second stage to inject subjects into target images.
Also, users can leverage subject injection methods~\cite{chen2024anydoor, chen2024zero} in the first stage and utilize text-guided editing methods~\cite{hertzprompt, deutch2024turboedit} for further subjects editing. 
We compare both two-stage approaches with our method and present results in Fig.~\ref{fig:compare_twostage}. As shown in the figure, Dreambooth with Paint-by-example is challenging to capture fine-grained features, leading to feature and identity inconsistency. MimicBrush with TurboEdit struggles to follow editing prompts, and the interaction with the background is not harmonious. Our approach achieves the best feature and identity preservation and adapts the generated results to the new scenario with prompt editing, surpassing the performance of two-stage methods.
\input{Figures/compare_twostage}

\subsection{Ablation study}

\paragraph{\textbf{Shift strength of $\alpha$.}}
In Eq.~\ref{eq:ours}, $\alpha_{1}$ and $\alpha_{2}$ are shift strength parameters for controlling guidance strength from textual prompts and subject images. We further examine the impact of different values for $\alpha_{1}$ and $\alpha_{2}$ on the generation results. When testing $\alpha_{1}$, $\alpha_{2}$ is set to 0.5, and vice versa.
The results, as shown in Fig.~\ref{fig:ablation_alpha}, indicate that as $\alpha_{1}$ increases, the expressiveness of the textual prompt in the generated result gradually strengthens. Similarly, increasing $\alpha_{2}$ enhances the image expression, causing the identity of the generated subject to align more closely with the reference. However, excessively large values of $\alpha_{1}$ and $\alpha_{2}$(e.g., 0.5) lead to a decrease in image quality. Therefore, users can adjust these parameters based on the specific image to balance the control of the image and generation quality.
\input{Figures/ablation_alpha}

\paragraph{\textbf{Customized subject insertion via basic Flux-Fill model}}
We build an inpainting pipeline with $I_c$ and $I_{s}$ concatenated as input to evaluate the basic customized subject insertion ability of Flux-Fill.
Results in Fig.~\ref{fig:main_compare} show that, to some extent, the basic pipeline generates subjects similar to $I_c$ (although the details are not aligned enough with the reference image); however, when editing the subjects, identity consistency is significantly reduced, indicating that without our methods, the basic model does not learn semantic level information, resulting in limited editing ability.

\input{Figures/ablation_blend}
\input{Figures/ablation_head}

\paragraph{\textbf{Without token blending}}
We ablate the token blending module and present results in Fig.~\ref{fig:ablation_blend}. We can see that with the introduction of latents from the target and fusion with latents from subjects along the denoising step, the presentation of subjects in the target images has better interaction with the background, achieving the overall consistency of the image (e.g., the subject and background hue in the image are consistent in the first case and the toy has better interaction with broom when editing it as sitting on the broom in the second case). We also present quantitative results in Tab.~\ref{tab:alignment}, and results show that without token blending, the FID score increases, indicating the overall image quality decreases. The CLIP-T score decreases, indicating that the results have a lower alignment with the prompts. Without token blending, subjects have less interaction with background, leading to less following the prompts.

\paragraph{\textbf{Without attention heads reweighting}}
We evaluate prompt representation by ablating attention heads reweighting method, reweighting all heads equally, and present results in Fig.~\ref{fig:ablation_head}. The findings reveal that, in the first case, due to the prior influence of images (white ceramic material), it is difficult to effectively edit the teapot (metal materials) without reweighting key heads. In the second case, although the text expresses ``white clothes'', the outputs are still affected by the reference and generate pink clothes, which fails to reflect the intended prompt.
Quantitative results in Tab.~\ref{tab:alignment} show that without reweighting, the FID score increases, indicating lower image quality. Furthermore, the decrease in CLIP-T score also shows reduced alignment with the editing prompt.
Overall, without head-wise enhancement for prompt representation, the generated results are greatly influenced by references and difficult to control by the prompt, leading to sub-optimal generation effects.

%% file: Figures/main_compare.tex
\begin{figure*}[!t]
  \centering
  \includegraphics[width=1\linewidth]{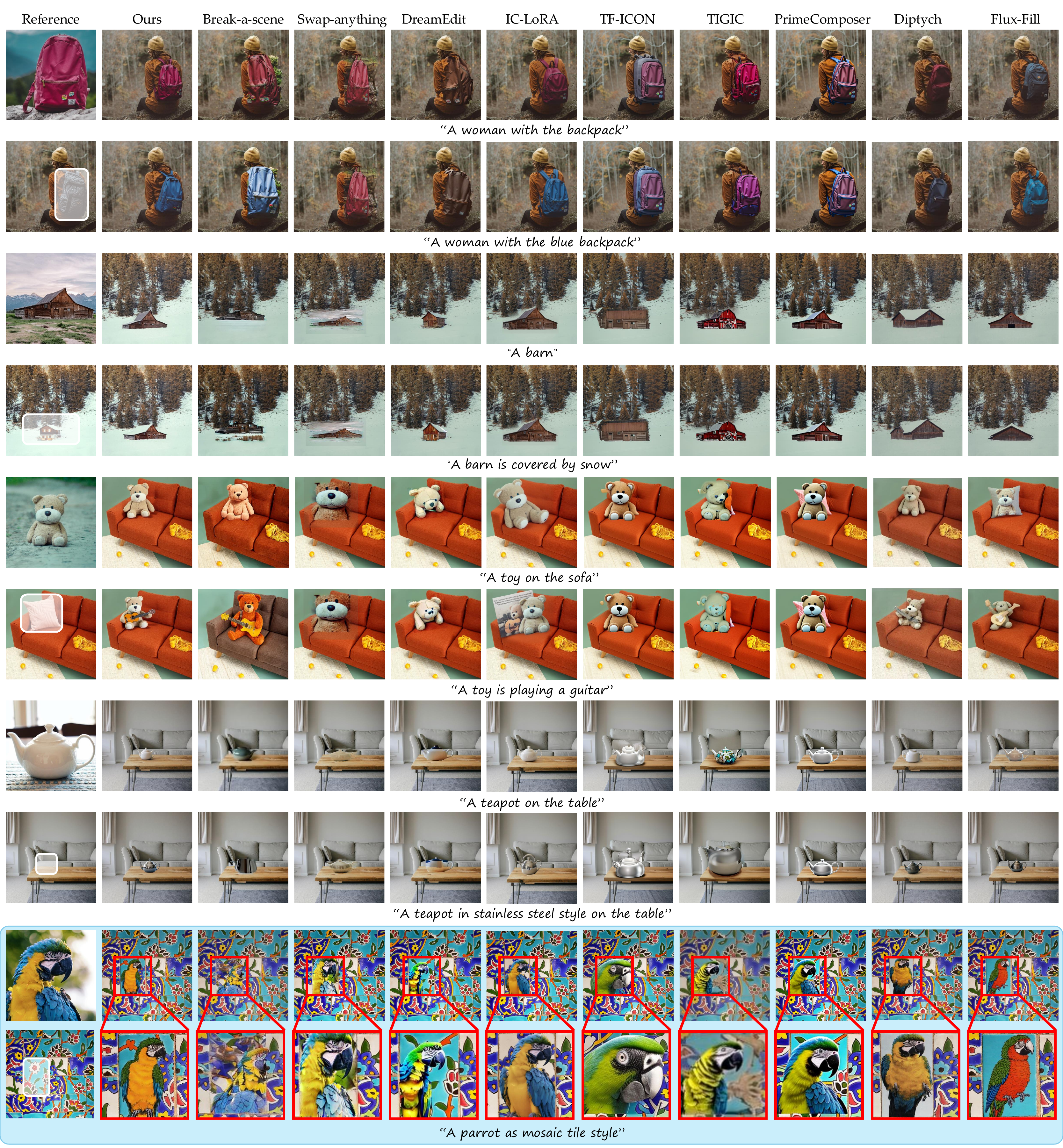}
  \caption{\textbf{Qualitative comparison on subject injection and editing with baseline methods.} Results of our results maintain identity consistency with reference while preserving fine-grained features, and are also aligning with the prompts. Masks are labeled as white boxes on target images.}
  \label{fig:main_compare}
\end{figure*}

%% file: Tables/alignment.tex
\begin{table*}[!th]
\centering
\caption{\textbf{Comparison of similarity scores between output images and reference images, and between output images and text prompts.}.  
``Injection'' evaluates the subject identity alignment between the reference images and the output images. ``Editing'' evaluates the text alignment between the output images and the corresponding prompts.
Our method has the best scores, indicating that our approach successfully edit images guided by text while maintaining consistency with reference images and high image quality.}
\vspace{-0.5em}
\label{tab:alignment}
\resizebox{\linewidth}{!}{%
\begin{tabular}{cccccccc}
\toprule
\multirow{2}{*}{\textbf{Methods}}&\multicolumn{2}{c}{\textbf{DINO}$(\uparrow)$} &\multicolumn{2}{c}{\textbf{CLIP-I}$(\uparrow)$}  &\multicolumn{2}{c}{\textbf{CLIP-T}$(\uparrow)$}        &    \multirow{2}{*}{\textbf{FID$(\downarrow)$}} \\ 
\cline{2-7} &\textbf{Injection} &\textbf{Editing} &\textbf{Injection} &\textbf{Editing} &\textbf{Injection} &\textbf{Editing}             \\ \hline
Break-a-scene   	&$0.7041 \pm 0.0659$ &$\textbf{0.7087} \pm 0.0546$	&$0.7128 \pm 0.2076$ &$0.6640 \pm 0.1855$	&$0.2570 \pm 0.0230$ &$0.2820 \pm 0.0659$ &217.81\\ 
Swap-anything   	&$0.7058 \pm 0.0625$	&$0.7035 \pm 0.0556$	&$0.7275 \pm 0.1821$	&$0.7296 \pm 0.1602$	&$0.2493 \pm 0.0181$	&$0.2296 \pm 0.0354$ &190.69\\  
DreamEdit           &$0.6521 \pm 0.0625$     &$0.6477 \pm 0.0652$   &$0.7203 \pm 0.1321$   &$0.7169 \pm 0.1377$     &$0.2295 \pm 0.0562$     &$0.2271 \pm 0.0557$ &$176.21$\\ 
IC-LoRA           &$0.7005\pm 0.0631$     &$0.6855 \pm 0.0621$   &$0.6891 \pm 0.1422$   &$0.6511 \pm 0.1325$     &$0.2523 \pm 0.0367$     &$0.2624 \pm 0.0522$ &$149.75$\\ \hline
DB+Paint-by-example    &$0.7037 \pm 0.0601$    &$0.6955 \pm 0.0442$    &$0.7162 \pm 0.0193$    &$0.6841 \pm 0.0167$    &$0.2668 \pm 0.0205$      &$0.2385 \pm 0.0343$  &$172.33$\\
MimicBrush+TurboEdit &$0.7044 \pm 0.0511$ &$0.6951 \pm 0.0502$ &$0.7107 \pm 0.1337$ &$0.6991 \pm 0.1851$ &$0.2674 \pm 0.0366$ &$0.2551 \pm 0.0621$ & 153.28 \\ \hline
TF-ICON & $0.7053 \pm 0.0280$ & $0.7061 \pm 0.0225$ & $0.7235 \pm 0.1188$ & $0.7018 \pm 0.1542$ & $0.2334 \pm 0.0270$ & $0.2234 \pm 0.0297$ &169.04\\
TIGIC & $0.6901 \pm 0.0231$ & $0.6906 \pm 0.0302$ & $0.7145 \pm 0.1744$ & $0.6789 \pm 0.2012$ & $0.2656 \pm 0.0313$ & $0.2272 \pm 0.0631$ &179.07\\
PrimeComposer & $0.7029 \pm 0.0510$ & $0.6931 \pm 0.0476$ & $0.7124 \pm 0.1128$ & $0.7426 \pm 0.0619$ & $0.2609 \pm 0.0163$ & $0.2300 \pm 0.0532$ &166.85\\
Diptych & $0.6559 \pm 0.0679$ & $0.6531 \pm 0.0691$ & $0.7225 \pm 0.1287$ & $0.7145 \pm 0.1140$ & $0.2321 \pm 0.0569$ & $0.2287 \pm 0.0403$ &179.28\\ \hline
Flux-Fill & $0.6798 \pm 0.0602$ & $0.6761 \pm 0.0497$ & $0.7668 \pm 0.1274$ & $0.7843 \pm 0.1124$ & $0.2634 \pm 0.0213$ & $0.2667 \pm 0.0288$ &127.18\\ 
Ours w/o head   &$0.7115 \pm 0.0564$ &$0.6891 \pm 0.0484$ &$0.7882 \pm 0.1201$ &$0.7889 \pm 0.1121$  &$0.2621 \pm 0.186$ &$0.2682 \pm 0.331$ &$123.82$ \\ 
Ours w/o blend  &$0.7116 \pm 0.0591$ &$0.6902 \pm 0.0519$ &$0.7901 \pm 0.1192$ &$0.7869 \pm 0.1133$  &$0.2633 \pm 0.171$ &$0.2703 \pm 0.294$ &$129.47$\\ \hline
Ours & $\textbf{0.7121} \pm 0.0542$ & $0.6945 \pm 0.0563$ & $\textbf{0.7957} \pm 0.1138$ & $\textbf{0.7924} \pm 0.1113$ & $\textbf{0.2685} \pm 0.0173$ & $\textbf{0.2834} \pm 0.0365$ &\textbf{122.61}\\ \bottomrule
\vspace{-2em}
\end{tabular}
}

\end{table*}

%% file: Figures/user_study.tex
\begin{figure}[t]
  \centering
  \includegraphics[width=1\linewidth]{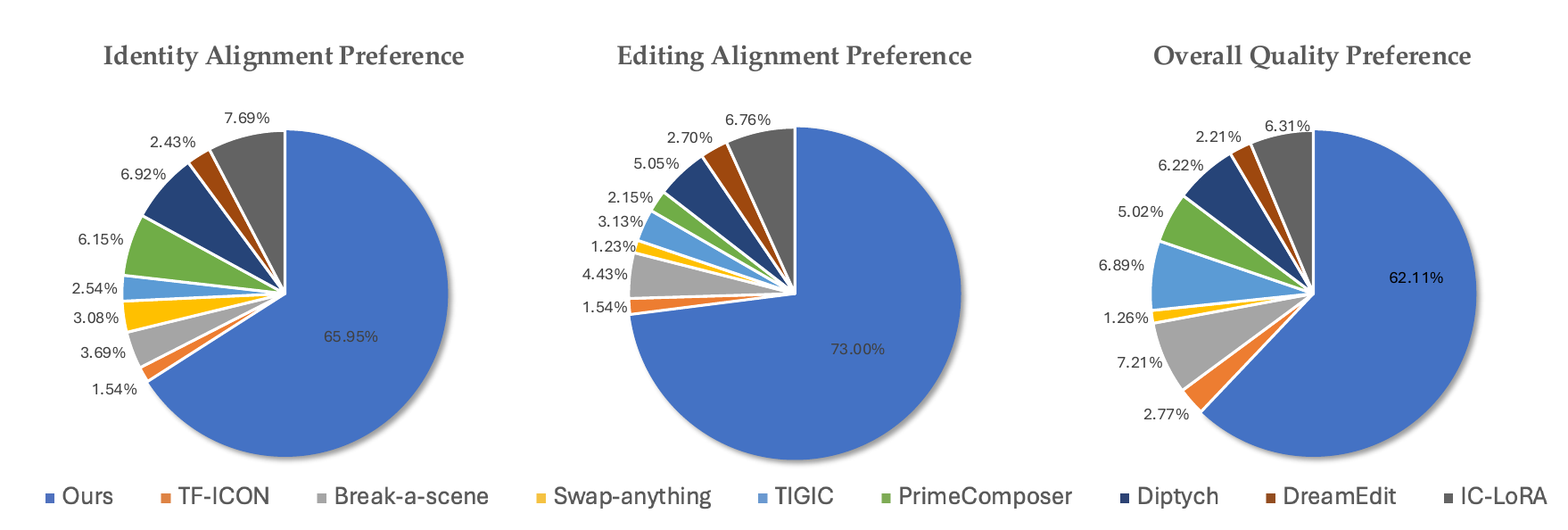}
  \caption{\textbf{User study results.} }
  \label{fig:user_study}
\end{figure}

%% file: Tables/alignment_foreback.tex
\begin{table}[!th]
\centering
\caption{\textbf{Quantitative comparison of the subject and background consistency with the subject and target images.} Higher scores in the ``Subject'' columns indicate better preservation of subject fidelity from the content images, while higher scores in the ``Background'' columns reflect better preservation of the target image background.}
\vspace{-0.5em}
\label{tab:alignment_foreback}
\scalebox{0.85}{
\begin{tabular}{ccccc}
\toprule
\multirow{2}{*}{\textbf{Methods}} & \multicolumn{2}{c}{\textbf{Subject}$(\uparrow)$} & \multicolumn{2}{c}{\textbf{Background}$(\uparrow)$} \\
\cline{2-5}
& \textbf{DINO} & \textbf{CLIP-I} & \textbf{DINO} & \textbf{CLIP-I} \\ \hline
Break-a-scene & 0.8415 & 0.6315 & 0.9203 & 0.7560 \\
Swap-anything & 0.8361 & 0.7329 & 0.9288 & 0.7673 \\
DreamEdit & 0.8504 & 0.7429 & 0.9585 & 0.7962 \\
IC-LoRA & 0.8179 & 0.7181 & 0.9333 & 0.7972 \\ \hline
DB+Paint-by-example &0.8224 &0.9385 &0.9305 &0.7851 \\
MimicBrush+TurboEdit &0.8301 &0.9297 &0.9342 &0.7921 \\ \hline
TF-ICON & 0.8450 & 0.7632 & 0.9160 & 0.7292 \\
TIGIC & 0.8483 & 0.6998 & 0.9428 & 0.7954 \\
PrimeComposer & 0.8505 & 0.7725 & 0.9405 & 0.7962 \\
Diptych & 0.7700 & 0.6560 & 0.8734 & 0.7570 \\ \hline
Flux-Fill & 0.8405 & 0.8001 & 0.9531 & 0.7853 \\ \hline
Ours & \textbf{0.8523} & \textbf{0.8090} & \textbf{0.9596} & \textbf{0.8100} \\ 
\bottomrule
\end{tabular}
}
\vspace{-1em}
\end{table}

%% file: Figures/compare_twostage.tex
\begin{figure}[htbp]
  \centering
  \includegraphics[width=0.9\linewidth]{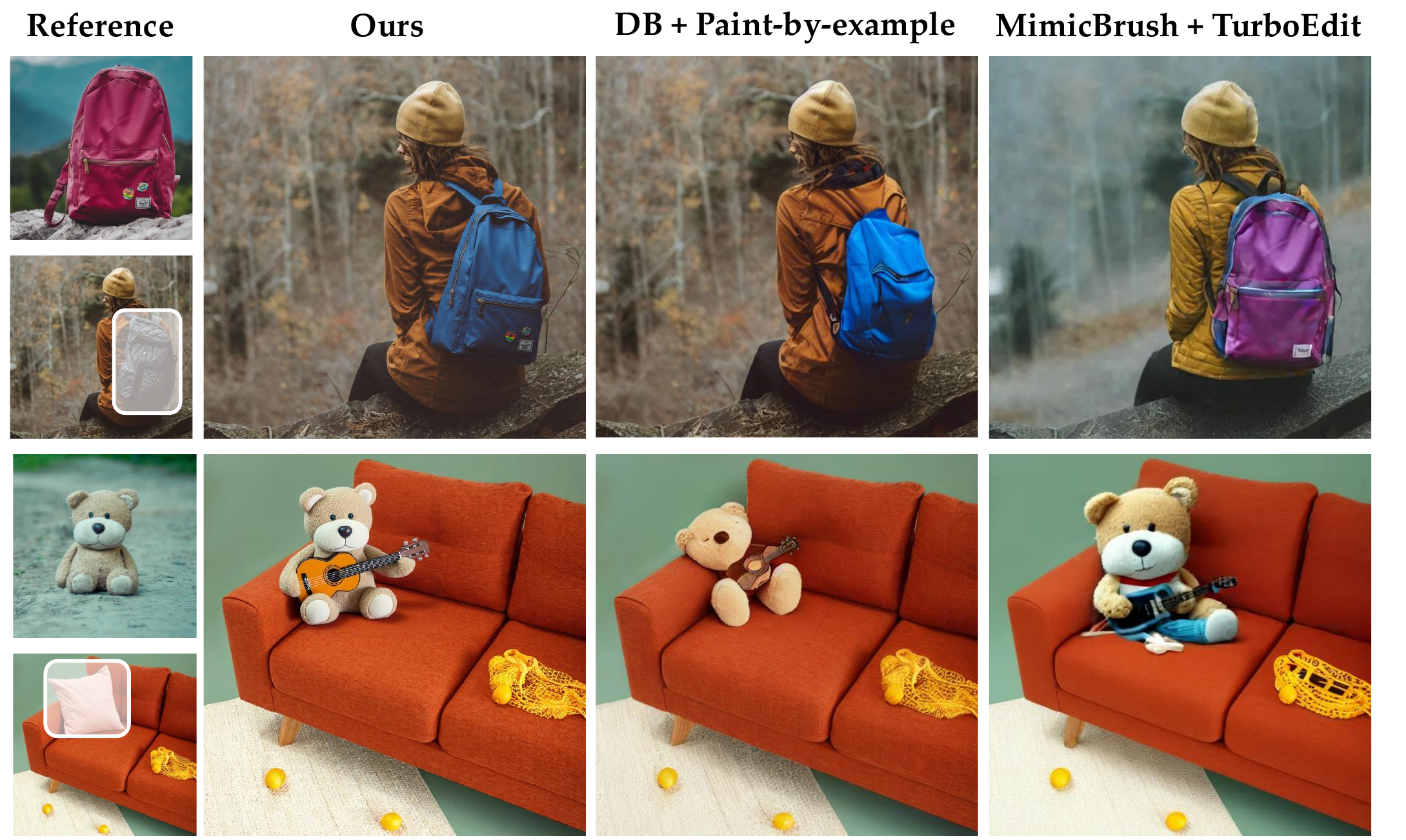}
  \caption{\textbf{Comparisons with two-stage methods.}}
  \label{fig:compare_twostage}
\end{figure}

%% file: Figures/ablation_alpha.tex
\begin{figure}[!h]
  \centering
  \includegraphics[width=0.9\linewidth]{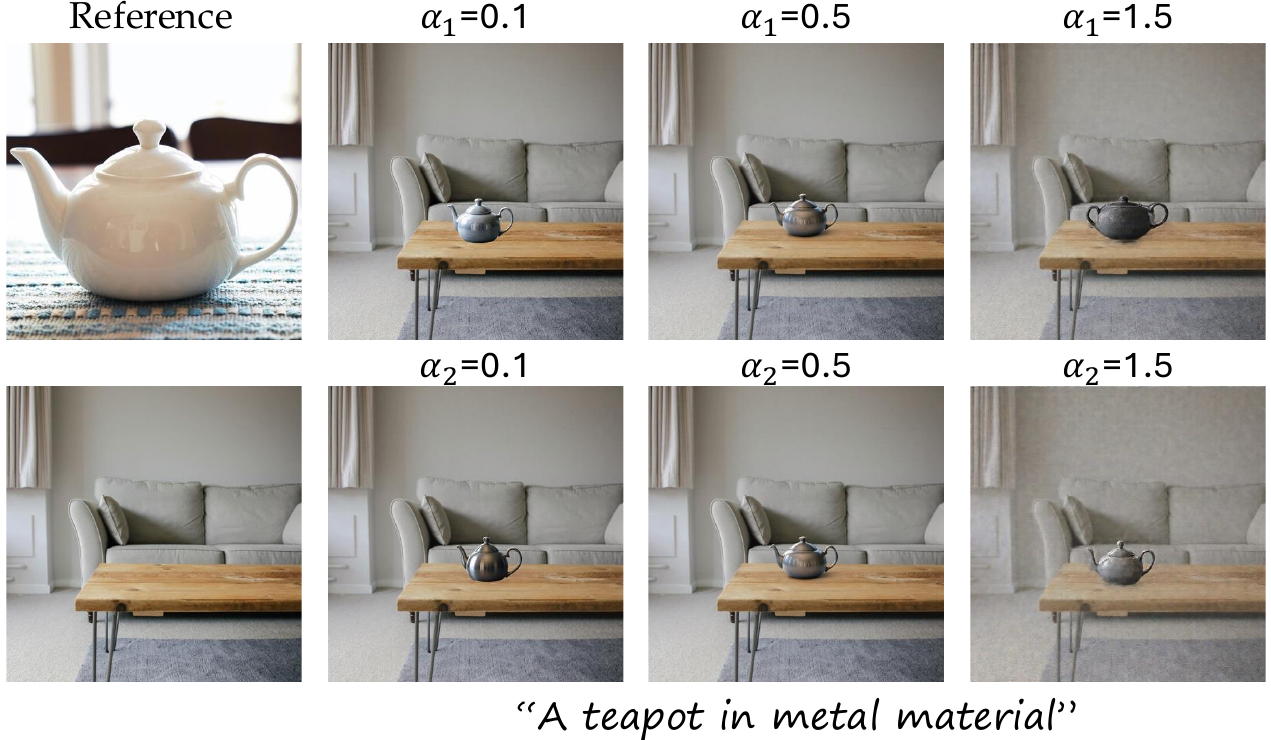}
  \caption{\textbf{Ablation study on shift strength.}}
  \label{fig:ablation_alpha}
\end{figure}

%% file: Figures/ablation_blend.tex
\begin{figure}[!h]
  \centering
  \includegraphics[width=0.9\linewidth]{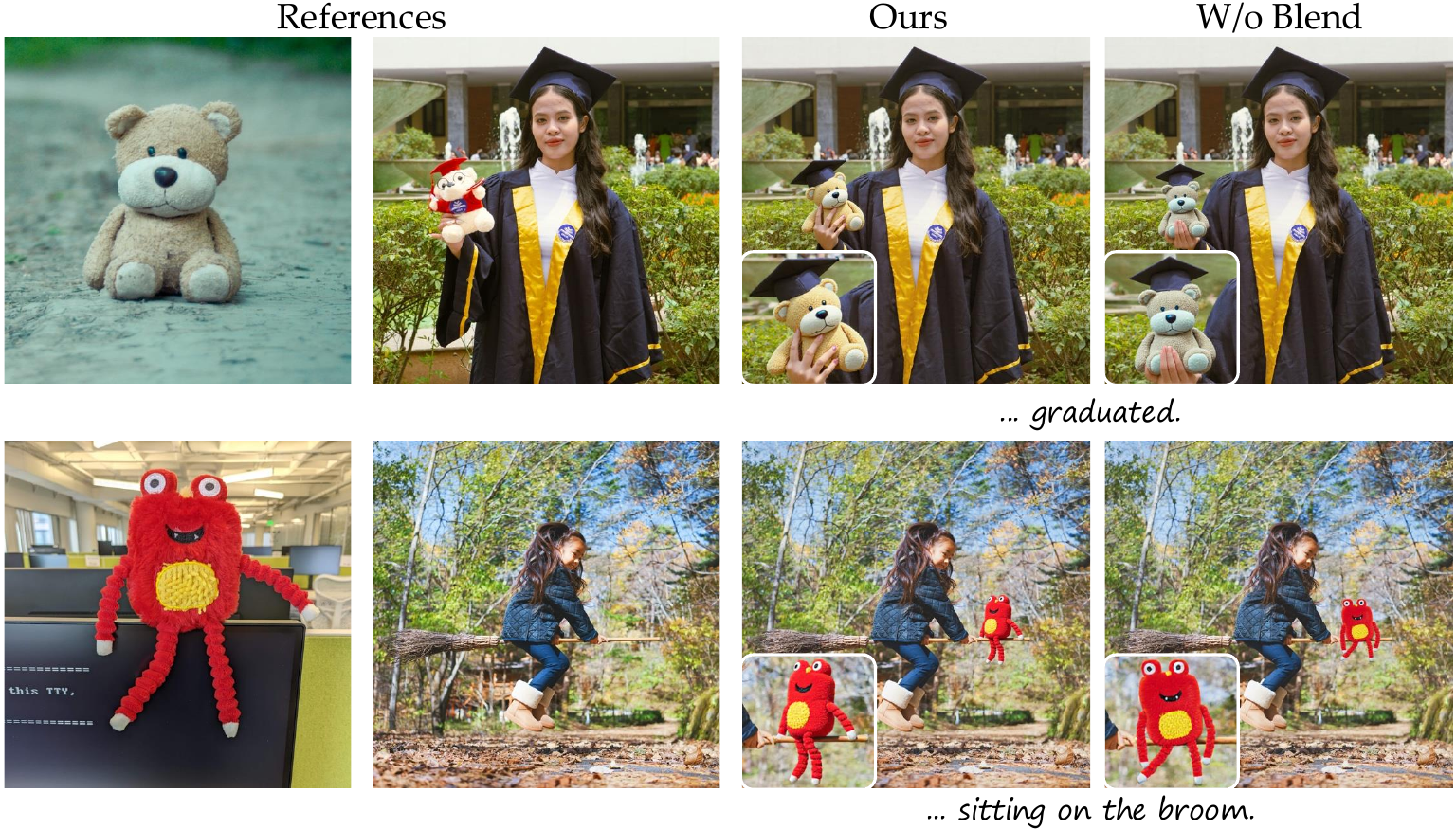}
  \caption{\textbf{Ablation study on token blending.}}
  \label{fig:ablation_blend}
\end{figure}


%% file: Figures/ablation_head.tex
\begin{figure}[!htbp]
  \centering
  \includegraphics[width=0.9\linewidth]{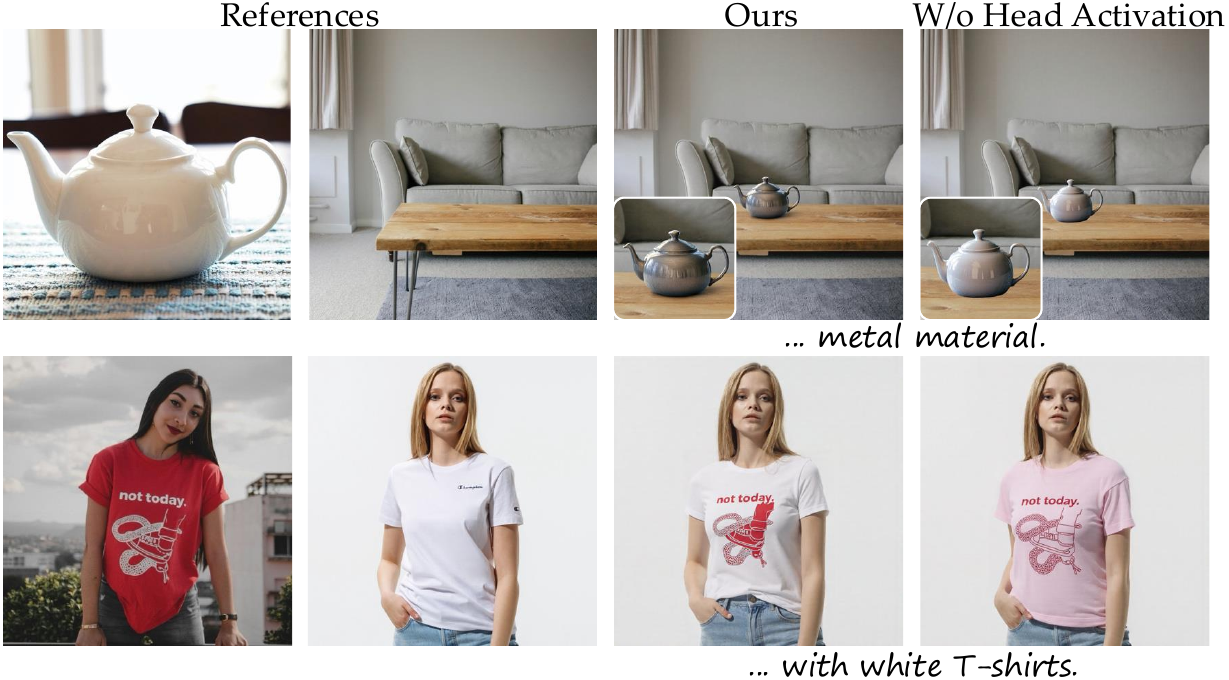}
  \caption{\textbf{Ablation study on attention heads activation.} }
  \label{fig:ablation_head}
\end{figure}

%% file: sec/6_app.tex
\section{Applications}
\paragraph{\textbf{Virtual try on.}}
A key application of our method is Virtual Try-On (VTON), which involves digitally dressing a target person with specified clothing. This is widely used in fashion retail to help users visualize outfits before purchase. As shown in Fig.~\ref{fig:application_vton}, our method accurately transfers garments to the target subject while preserving identity and achieving strong alignment with user prompts. It handles various clothing types and styles, demonstrating versatility for real-world fashion scenarios.
\input{Figures/application_vton}

\paragraph{\textbf{Compositional generation.}}
Another application of our method is compositional generation, where users iteratively insert multiple elements into a scene with layout control via masks. This is especially useful in tasks like interior design, enabling users to explore combinations of furniture by adjusting placements and styles. As shown in Fig.~\ref{fig:application_compo}, our framework supports flexible and coherent scene construction, allowing users to refine designs interactively and visualize personalized arrangements with ease.
\input{Figures/application_compo}

\input{Figures/application_part}
\paragraph{\textbf{Partly insertion.}}
Our method supports selective part-based insertion, enabling users to transfer specific regions from reference into corresponding locations of the generated output. This allows fine-grained control while maintaining spatial and semantic alignment. As shown in Fig.~\ref{fig:application_part}, we insert the wheels of a reference car and the legs of chairs into target scenes. In both cases, the inserted parts preserve high fidelity and blend naturally with the background, demonstrating the method’s effectiveness for precise partial edits in applications such as product variation and scene refinement.

%% file: Figures/application_vton.tex
\begin{figure}[!htb]
  \centering
  \includegraphics[width=0.9\linewidth]{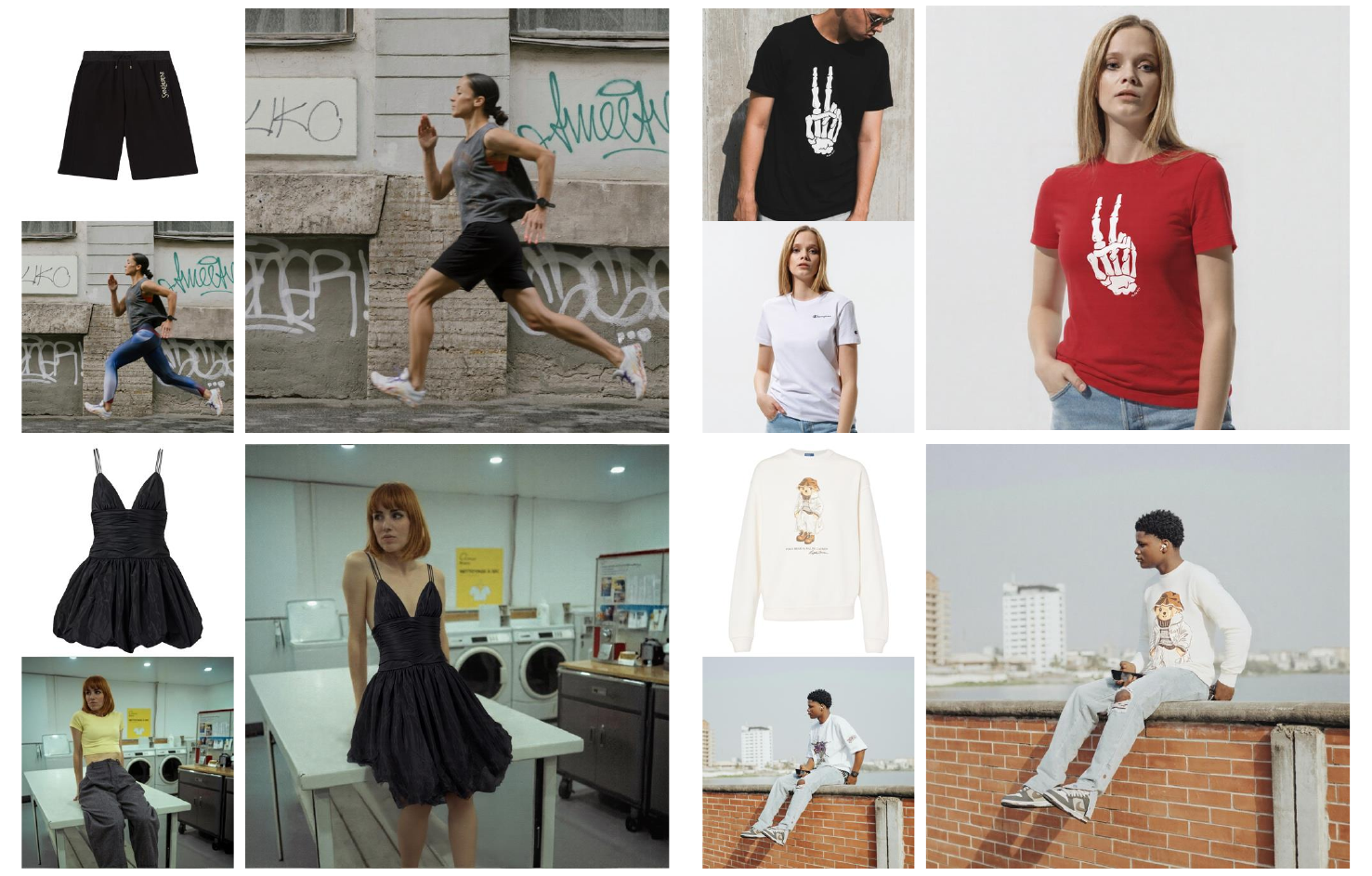}
  \caption{\textbf{Application of virtual try-on.} }
  
  \label{fig:application_vton}
\end{figure}

%% file: Figures/application_compo.tex
\begin{figure}[!htbp]
  \centering
  \includegraphics[width=0.9\linewidth]{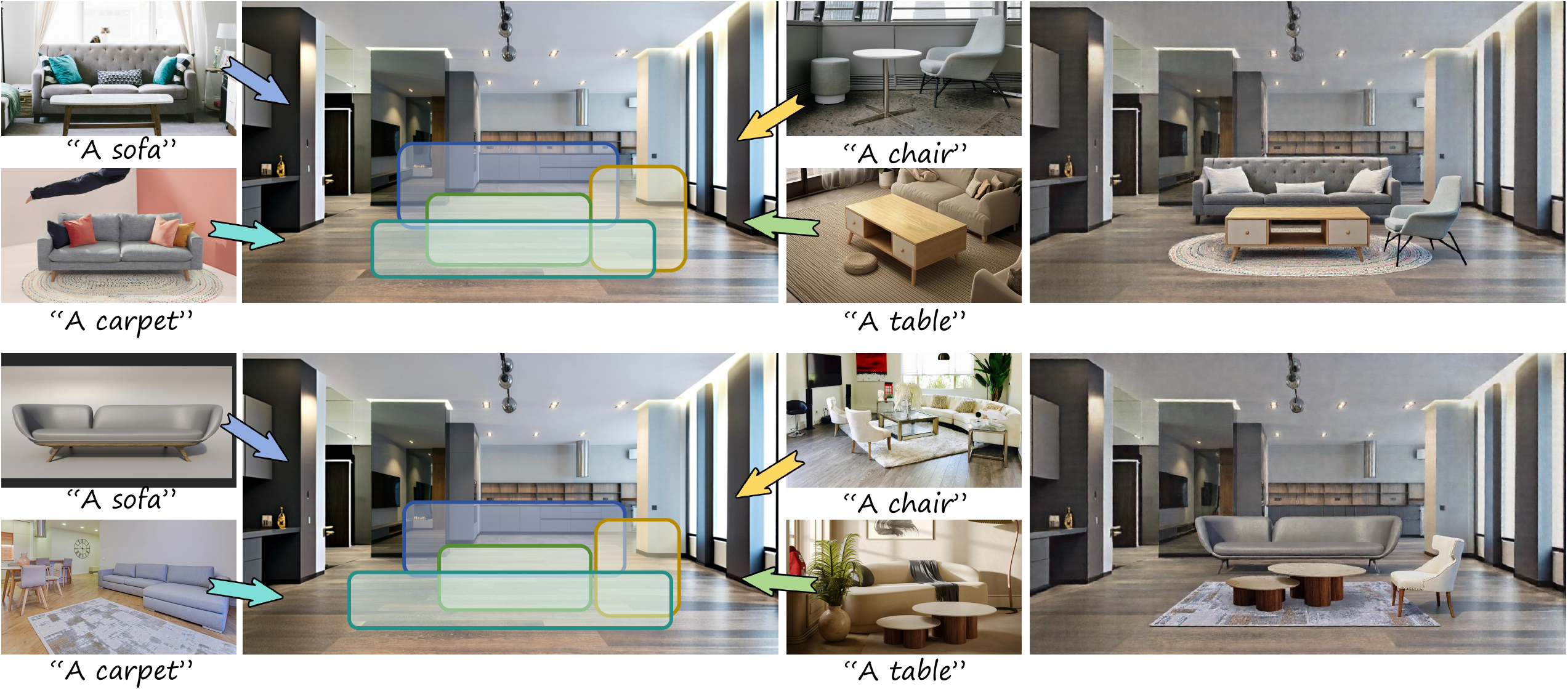}
  \caption{\textbf{Application of compositional generation.} }
  \label{fig:application_compo}
\end{figure}

%% file: Figures/application_part.tex
\begin{figure}[!htb]
  \centering
  \includegraphics[width=0.75\linewidth]{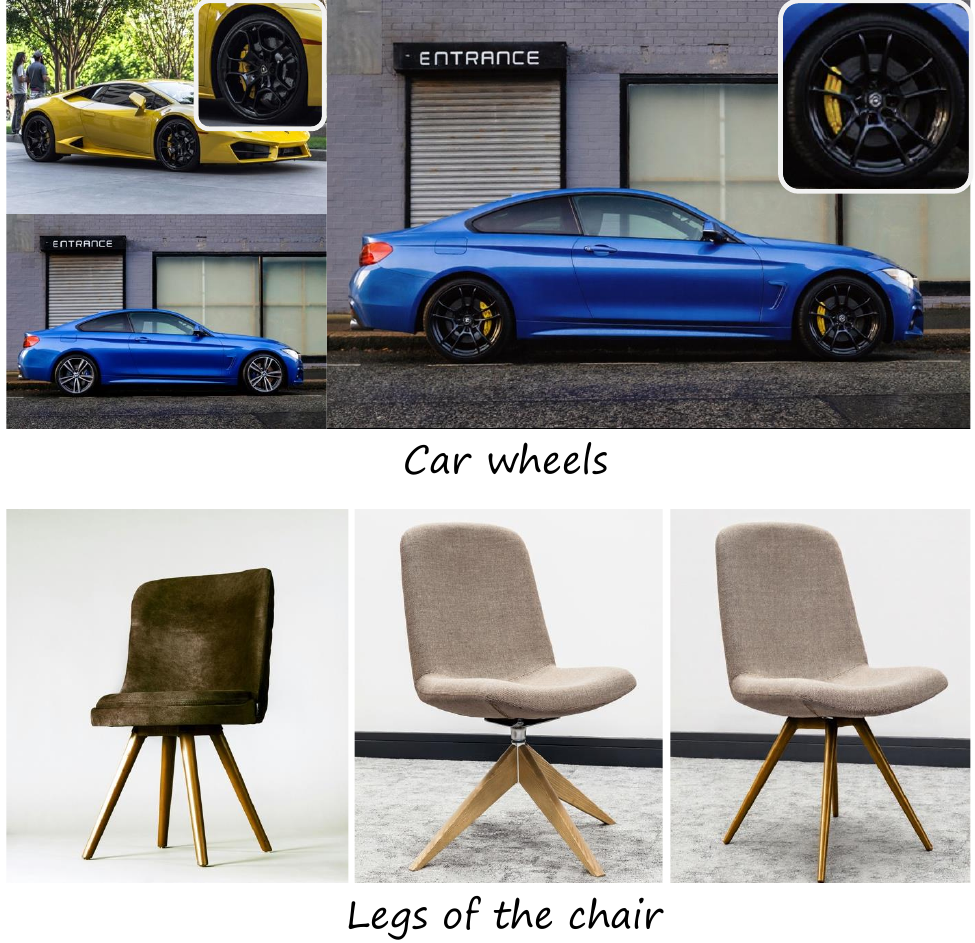}
  \caption{\textbf{Application of partly insertion.} }
  \label{fig:application_part}
\end{figure}

%% file: sec/7_limit_and_conclusion.tex
\section{Limitations and badcase}
When target images contain subjects similar to those in the subject images, the generated results may exhibit features resembling the target image. For example, as shown in Fig.~\ref{fig:badcase}, certain patterns on the windows of the generated vehicle are similar to the corresponding positions in the target image’s background. This issue likely arises because similar contextual features are referenced during the self-attention calculation.
To address this in future work, we can consider introducing constraints on the attention mechanism.
\input{Figures/badcase}

\section{Conclusions}
In this work, we leverage ICL to activate the context-consistent generation capability of large-scale pre-trained text-to-image models, enabling customized subject insertion. By reformulating ICL as latent space shifting, we achieve zero-shot insertion of specific subjects into novel images. Additionally, we employ head-wise reweighting and token blending to enhance the insertion consistency of text attribute expression. Extensive quantitative and qualitative experiments and user study demonstrate the superiority of our approach over existing state-of-the-art methods.

%% file: Figures/badcase.tex
\begin{figure}[!htbp]
  \centering
  \includegraphics[width=0.95\linewidth]{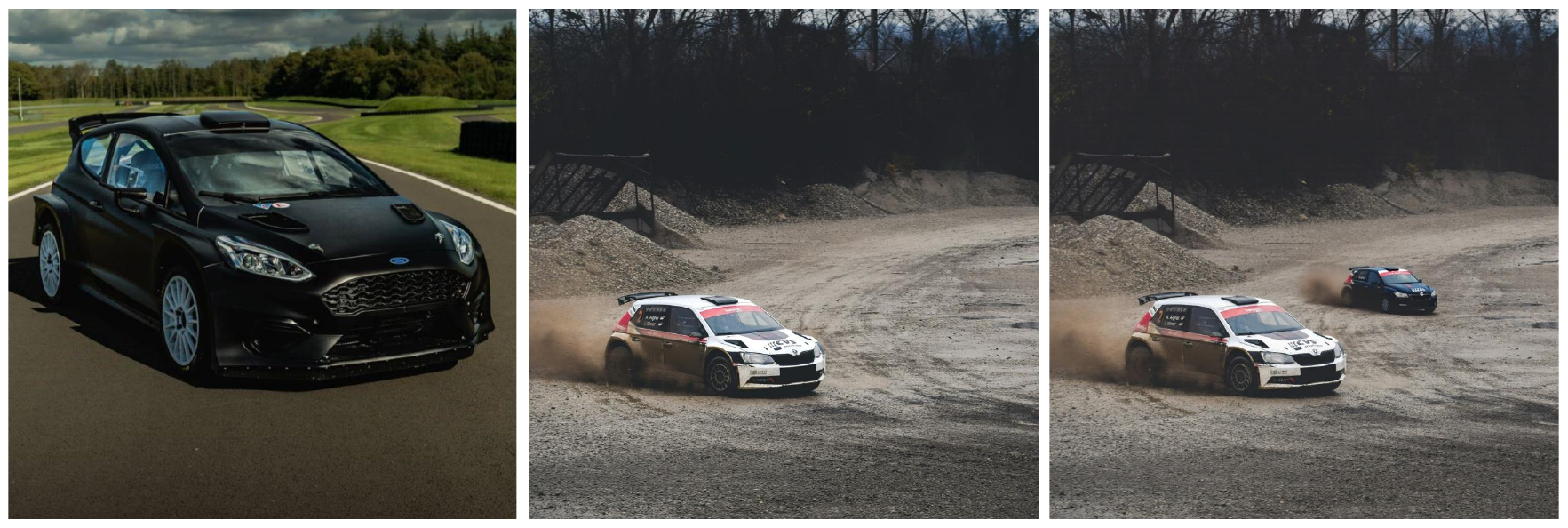}
  \caption{\textbf{Badcase.} In some cases, interaction among similar contextual features in attention calculation may cause same features in appearance of results as the concepts from background.}
  \label{fig:badcase}
\end{figure}

%% file: sup.tex
\pagestyle{plain} 


\renewcommand{\algorithmcfname}{ALGORITHM}
\SetAlFnt{\small}
\SetAlCapFnt{\small}
\SetAlCapNameFnt{\small}
\SetAlCapHSkip{0pt}

\newcommand{\hide}[1]{}
\newcommand{\od}[1]{{\color{red}#1}}

 
 \begin{center}
    {\LARGE \bf Supplementary Materials \\[0.5em]}
\end{center}

\maketitle

\balance
\section{Derivation of Joint-attention Mechanism}\label{formula}
To derive the formulation for the joint-attention mechanism of MM-DiTs, we represent the input embedding by $X = \operatorname{Concatenate}(\left[x_{p}, x_{c}, x_{s}\right])$, where $x_{p}$, $x_{c}$ and $x_{s}$ represent input token embeddings at the same concatenating positions as $p$, $I_{c}$ and $I_{s}$, respectively.
Let $W_q$, $W_k$, and $W_v$ be the learnable key, query, and value matrices for computing the attention features $Q$, $K$, and $V$, the output hidden states of attention blocks can be formulated as:
\small{
\begin{equation}
  \begin{aligned} 
  \hat{h}
  &=\operatorname{Attn}\left(Q, K, V\right)\\
  &=\operatorname{Attn}\left(X W_q, X W_k, X W_v\right)\\
  &=\operatorname{Attn}\left(\begin{bmatrix}
    x_p \\
    x_{c} \\
    x_{s}
    \end{bmatrix} W_q, \begin{bmatrix}
    x_p \\
    x_{c} \\
    x_{s}
    \end{bmatrix} W_k, \begin{bmatrix}
    x_p \\
    x_{c} \\
    x_{s}
    \end{bmatrix} W_v \right), \\
  &=\operatorname{Softmax}\left(\begin{bmatrix}
    x_p W_{qk} x_p^{\top} & x_p W_{qk} x_{c}^{\top} & x_p W_{qk} x_{s}^{\top} \\
    x_{c} W_{qk} x_p^{\top} & x_{c} W_{qk} x_{c}^{\top} & x_{c} W_{qk} x_{s}^{\top} \\
    x_{s} W_{qk} x_p^{\top} & x_{s} W_{qk} x_{c}^{\top} & x_{s} W_{qk} x_{s}^{\top}
    \end{bmatrix}\right) \begin{bmatrix}
    x_p W_v \\
    x_{c} W_v \\
    x_{s} W_v
    \end{bmatrix},\\
  &=\begin{bmatrix}
    A_{p,p} & A_{p,c} & A_{p,s} \\
    A_{c,p} & A_{c,c} & A_{c,s} \\
    A_{s,p} & A_{s,c} & A_{s,s}
    \end{bmatrix} \begin{bmatrix}
    x_p W_v \\
    x_{c} W_v \\
    x_{s} W_v
    \end{bmatrix},\\
  &=\operatorname{Concatente}(\left[h_{p}, h_{c}, h_{s}\right]),
  \end{aligned}
  \label{eq:attention}
\end{equation}
}
where $W_{qk} = W_{q} W_{k}^{\top}$, and $h_{p}$, $h_{c}$, $h_{s}$ represent the hidden states corresponding to each component in $X$. 
In addition, we use $A_{i,j}$ to represent the attention map in the position of patch $x_i W_{qk} x_j^{\top}$.

\section{Derivation for Equation 3 in Main Text}\label{equation3}
Since the generated result $I_{gen}$ is directly related to $h_{s}$, we can only focus on the last line of Eq.~\ref{eq:attention} and rewrite it in the form of the attention operation:
\small{
\begin{equation}
  \begin{aligned} 
  &h_{s}=\operatorname{Softmax}\left(\begin{bmatrix}
    x_{s} W_{qk} x_p^{\top} & x_{s} W_{qk} x_{c}^{\top} & x_{s} W_{qk} x_{s}^{\top}
    \end{bmatrix}\right) \begin{bmatrix}
    x_p W_v \\
    x_{c} W_v \\
    x_{s} W_v
    \end{bmatrix} \\
  &=\alpha_{p} \cdot \operatorname{Attn}\left(x_{s} W_q, x_{p} W_k, x_{p} W_v\right) 
  + \alpha_{c} \cdot \operatorname{Attn}\left(x_{s} W_q, x_{c} W_k, x_{c} W_v\right) \\
    \end{aligned}
\end{equation}
 \begin{equation}
  \begin{aligned}  
  & +\alpha_{s} \cdot \operatorname{Attn}\left(x_{s} W_q, x_{s} W_k, x_{s} W_v\right) \\
  &= \alpha_{p} \cdot h(prompt) + \alpha_{c} \cdot h(subject) + \alpha_{s} \cdot h(output) \\
  &= \alpha_{p} \cdot h(demo\_p) + \alpha_{c} \cdot h(demo\_c) + \alpha_{s} \cdot h(query),
  \end{aligned}
  \label{eq:final}
\end{equation}
}
where $\alpha_{p} + \alpha_{c} + \alpha_{s} =1$.

\input{sup/Tables/sup_alignment}
\input{sup/Tables/time}

\section{Implementation Details}
We employ Flux-1.0-fill[dev] with default hyperparameters as the base model. All baseline approaches follow their official implementations, with hyperparameters set accordingly.
For training-based methods, we utilize DreamBooth to learn the custom subject. Specifically:
\begin{itemize}
    \item For the Break-A-Scene baseline, we first learn custom subject and scene separately for both 800 steps using different placeholder words, then combine both in a prompt for joint generation.
    \item For the Swap-Anything baseline, we apply null-text inversion via DDIM to invert background images. Grounding DINO and Segment Anything are used for object detection and mask extraction. During the swapping process, the steps for latent image feature, cross-attention map, self-attention map, and self-attention output are set to 30, 20, 25, respectively.
    \item For the DreamEdit baseline, Segment Anything is used to obtain the mask of the subject. The number of iterations is set to five for the replacement task and seven for the addition task. The mask dilation kernel is set to 20. The encoding ratio is set to be 0.8 for the first iteration and decreases linearly as $k_{i} / T = k_{1} / T - i * 0.1$.
\end{itemize}

For training-free methods, TF-ICON, TIGIC, and PrimeComposer all use DPM-Solver++ for image inversion:
\begin{itemize}
    \item TF-ICON: Since both subject and background images belong to the photorealism domain, we set the classifier-free guidance (CFG) scale to 2.5. The threshold for injecting composite self-attention maps is set to 0.4, while the background preservation threshold is 0.8.
    \item TIGIC: The CFG scale is set to 5, the composite self-attention injection threshold to 0.5, and the background preservation threshold to 0.8.
    \item PrimeComposer: The CFG scale is 2.5, and the hyperparameter for prior weight infusion is 0.2.
\end{itemize}

We utilize the implementation available on GitHub~\footnote{\url{https://github.com/wuyou22s/Diptych}}. The attention reweighting coefficient is set to 1.3.

\section{Robustness to Random Seeds}
All baseline comparisons used identical random seeds for fair evaluation. We conduct additional experiments using ten different random seeds to evaluate the stability of our method. The results in Tab.~\ref{tab:sup_alignment} show minimal variance across seeds, with average performance metrics closely aligning with those reported in Tab.1 of the main text. This confirms the robustness and consistency of our approach across different initializations.




%% file: sup/Tables/sup_alignment.tex
\begin{table*}[!h]
\centering
\caption{\textbf{Quantitative comparison of our method and baseline approaches \underline{across multiple random seeds}.} To evaluate the stability and consistency of the generated results, we conduct our method and all baseline methods using the same set of random seeds. The table demonstrates the consistency and robustness of our method across varying initialization conditions.}
\vspace{-0.5em}
\label{tab:sup_alignment}
\resizebox{\linewidth}{!}{%
\begin{tabular}{cccccccc}
\toprule
\multirow{2}{*}{\textbf{Methods}}&\multicolumn{2}{c}{\textbf{DINO}$(\uparrow)$} &\multicolumn{2}{c}{\textbf{CLIP-I}$(\uparrow)$}  &\multicolumn{2}{c}{\textbf{CLIP-T}$(\uparrow)$}        &    \multirow{2}{*}{\textbf{FID$(\downarrow)$}} \\ 
\cline{2-7} &\textbf{Injection} &\textbf{Editing} &\textbf{Injection} &\textbf{Editing} &\textbf{Injection} &\textbf{Editing}             \\ \hline
Break-a-scene  &$0.7038 \pm 0.0677$ &$\textbf{0.7088} \pm 0.0542$	&$0.7131 \pm 0.2081$ &$0.6642 \pm 0.1841$	&$0.2565 \pm 0.0239$ &$0.2822 \pm 0.0644$ &217.83\\ 
Swap-anything  &$0.7056 \pm 0.0634$	&$0.7032 \pm 0.0551$	&$0.7277 \pm 0.1816$  &$0.7291 \pm 0.1611$&$0.2494 \pm 0.0177$	&$0.2291 \pm 0.0343$ &190.72\\  \hline
DreamEdit           &$0.6524 \pm 0.0626$     &$0.6474 \pm 0.0655$   &$0.7206 \pm 0.1319$   &$0.7168 \pm 0.1375$     &$0.2297 \pm 0.0564$     &$0.2273 \pm 0.0559$ &$176.22$\\ 
IC-LoRA           &$0.7001\pm 0.0632$     &$0.6857 \pm 0.0622$   &$0.6894 \pm 0.1423$   &$0.6513 \pm 0.1327$     &$0.2518 \pm 0.0366$     &$0.2626 \pm 0.0521$ &$149.73$\\ \hline
DB+Paint-by-example    &$0.7036 \pm 0.0602$    &$0.6956 \pm 0.0441$    &$0.7161 \pm 0.0193$    &$0.6845 \pm 0.0166$    &$0.2665 \pm 0.0202$      &$0.2382 \pm 0.0341$  &$172.35$\\
MimicBrush+TurboEdit &$0.7043 \pm 0.0512$ &$0.6952 \pm 0.0504$ &$0.7105 \pm 0.1338$ &$0.6992 \pm 0.1852$ &$0.2675 \pm 0.0365$ &$0.2553 \pm 0.0624$ & 153.27 \\ \hline
TF-ICON & $0.7054 \pm 0.0275$ & $0.7064 \pm 0.0227$ & $0.7236 \pm 0.1193$ & $0.7015 \pm 0.1546$ & $0.2336 \pm 0.0265$ & $0.2237 \pm 0.0292$ &168.98\\
TIGIC & $0.6901 \pm 0.0231$ & $0.6906 \pm 0.0302$ & $0.7145 \pm 0.1744$ & $0.6789 \pm 0.2012$ & $0.2656 \pm 0.0313$ & $0.2272 \pm 0.0631$ &179.07\\
PrimeComposer & $0.7027 \pm 0.0513$ & $0.6933 \pm 0.0479$ & $0.7127 \pm 0.1133$ & $0.7424 \pm 0.0612$ & $0.2605 \pm 0.0158$ & $0.2227 \pm 0.0547$ &166.89\\
Diptych & $0.6555 \pm 0.0684$ & $0.6532 \pm 0.0685$ & $0.7228 \pm 0.1291$ & $0.7148 \pm 0.1134$ & $0.2318 \pm 0.0561$ & $0.2282 \pm 0.0412$ &179.32\\ 
Flux-Fill & $0.6801 \pm 0.0698$ & $0.6757 \pm 0.0499$ & $0.7665 \pm 0.1271$ & $0.7841 \pm 0.1127$ & $0.2637 \pm 0.0211$ & $0.2669 \pm 0.0285$ &127.16\\ \hline
Ours & $\textbf{0.7123} \pm 0.0539$ & $0.6944 \pm 0.0567$ & $\textbf{0.7959} \pm 0.1134$ & $\textbf{0.7928} \pm 0.1109$ & $\textbf{0.2689} \pm 0.0177$ & $\textbf{0.2836} \pm 0.0369$ &\textbf{122.63}\\ \bottomrule
\vspace{-2em}
\end{tabular}
}

\end{table*}

%% file: sup/Tables/time.tex
\begin{table}[!th]
\centering
\caption{\textbf{Comparison of time efficiency with baseline methods}.}
\vspace{-0.5em}
\label{tab:time}
\resizebox{0.8\linewidth}{!}{%
\begin{tabular}{cccccccc}
\toprule
\textbf{Methods} & \textbf{Training Time} & \textbf{Inference Time} \\ \hline
Break-a-scene &  328s & 15s \\ 
Swap-anything & 840s & 71s\\
DreamEdit & 1440s &35s\\ \hline
DB+Paint-by-example &845s &7s\\
MimicBrush+TurboEdit & - & 38s\\ \hline
TF-ICON & - & 49s\\
TIGIC & - & 48s \\
PrimeComposer & - & 41s\\ 
Diptych & - & 41s\\ \hline
Flux-Fill & - &48s\\
Ours & - & 52s \\ 
\bottomrule
\vspace{-0.5em}
\end{tabular}
}
\end{table}